\documentclass[letterpaper]{article}

\usepackage{mathtools}
\usepackage{booktabs}
\usepackage{latexsym} 
\usepackage[usenames,dvipsnames]{color}
\usepackage{amsmath}
\usepackage{amssymb}

\usepackage{color}
\def\year{2018}\relax
\usepackage{aaai18}  
\usepackage{times}  
\usepackage{helvet}  
\usepackage{courier}  
\usepackage{url}  
\usepackage{graphicx}  
\nocopyright
\usepackage{units}
\usepackage{algorithm}
\usepackage{algpseudocode}
\usepackage[usenames,dvipsnames,svgnames,table]{xcolor}
\usepackage{todonotes}

\newtheorem{df}{Definition}

\newtheorem{theorem}{Theorem}
\newtheorem{lemma}{Lemma}
\newtheorem{col}{Corollary}
\newcommand{\bt}{\begin{theorem}\em}
	\newcommand{\et}{\end{theorem}}
\newcommand{\bl}{\begin{lemma}\em}
	\newcommand{\el}{\end{lemma}}
\newcommand{\bc}{\begin{col}\em}
	\newcommand{\ec}{\end{col}}

\newcommand{\bea}{\begin{eqnarray}}
\newcommand{\eea}{\end{eqnarray}}

\newcommand{\bdf}{\begin{df}\em}
	\newcommand{\edf}{\end{df}}

\newcommand{\ben}{\begin{enumerate}}
	\newcommand{\een}{\end{enumerate}}

\newcommand{\argmax}{\operatornamewithlimits{argmax}}

\newcommand{\citea}[1]
{\citeauthor{#1} (\citeyear{#1})}

\usepackage{caption}
\usepackage{subcaption}

\algdef{SE}[VARIABLES]{Variables}{EndVariables}
{\algorithmicvariables}
{\algorithmicend\ \algorithmicvariables}
\algnewcommand{\algorithmicvariables}{\textbf{global variables}}

\usepackage[normalem]{ulem}

\newcommand{\citet}[1]{\citeauthor{#1}~\shortcite{#1}}

\newboolean{includeMemo}
\setboolean{includeMemo}{true} 

\newcommand{\memoside}[1]{\ifthenelse{\boolean{includeMemo}}{\todo[caption={},color=green!20!]{{\footnotesize #1}}}}
\newcommand{\memo}[1]{\ifthenelse{\boolean{includeMemo}}{\todo[inline,caption={},color=green!20!]{#1}}}
\title{Automatic Algorithm Selection in Multi-Agent Pathfinding}
\author{Devon Sigurdson \\ {\bf\Large Vadim Bulitko} \\ University of Alberta \\ {dbsigurd, bulitko@ualberta.ca} \And
 William Yeoh \\ 	Washington University in St. Louis \\ wyeoh@wustl.edu \And 
 Carlos Hern{\'a}ndez \\ Universidad Andres Bello \\ carlos.hernandez.u@unab.cl \And
 Sven Koenig \\ University of Southern California \\ 	skoenig@usc.edu
}
\begin{document}

	\maketitle

	\begin{abstract}

In a multi-agent pathfinding (MAPF) problem, agents need to navigate from their start to their goal locations without colliding into each other. There are various MAPF algorithms, including Windowed Hierarchical Cooperative A*, Flow Annotated Replanning, and Bounded Multi-Agent A*. It is often the case that there is no single algorithm that dominates all MAPF instances. Therefore, in this paper, we investigate the use of deep learning to automatically select the best MAPF algorithm from a portfolio of algorithms for a given MAPF problem instance. Empirical results show that our automatic algorithm selection approach, which uses an off-the-shelf convolutional neural network, is able to outperform any individual MAPF algorithm in our portfolio.

	\end{abstract}
	
	\section{Introduction}
	
Pathfinding is a common task for many applications, such as robotics, transportation, and video games. Often, paths need to be planned for multiple game characters, for example, when each one of several game characters needs to move from its current location to a given goal location. Single-agent pathfinding is a simpler problem since shortest single-agent paths can be found optimally in polynomial time with search methods like A*~\cite{AStar}. Multi-agent pathfinding (MAPF) is a more complex problem because one needs to avoid collisions among the agents. Artificial intelligence and robotics have developed a large number of MAPF algorithms. Suboptimal complete MAPF algorithms often implement specific movement rules for agents that guarantee completeness and run fast since they avoid search. An example is Push and Swap/Rotate~\cite{Luna:11,Wild:13}. Some optimal or bounded-suboptimal MAPF algorithms reduce the MAPF problem to other combinatorial problems. Examples include reductions to CSP \cite{Ryan:10}, SAT~\cite{Sury:12}, ILP~\cite{Yu:13c}, and ASP~\cite{Erde:13}. Others are based on heuristic search. Examples are M*~\cite{Wagn:15}, Conflict-Based Search~\cite{Shar:15,Boya:15,Cohe:15} and many others. A longer recent overview of MAPF algorithms is provided by~\citea{Feln:17}.

While finding collision-free paths for multiple agents can be done in polynomial
time~\cite{korn1984}, in practice, these paths should also be reasonably short.
Finding collision-free paths that optimize solution quality measured as
makespan (the largest arrival time of any agent at its goal location) or
flowtime (the sum of the arrival times of all agents at their goal locations)
is NP-hard~\cite{Yu:13a}. In some cases, even approximating the optimal
solution quality is NP-hard~\cite{Ma:16a}. Unfortunately, paths with a
reasonably good solution quality often need to be found quickly (e.g., when planning paths for game characters online). Researchers in artificial intelligence have developed a variety of MAPF algorithms that can be used for
this purpose, such as Windowed Hierarchical A*~\cite{silver:05}, Flow Annotated Replanning~\cite{Wang:2008} and Bounded Multi-agent A*~\cite{Sigurdson2018}. To the best of our knowledge, none of the MAPF algorithms universally dominates all others. 

As a starting point of determining a good MAPF algorithm, we are primarily interested in ensuring that all agents are successful, in the sense that they reach their goal locations by a given deadline. Unfortunately, it has recently been shown to be NP-hard to determine whether all agents can be successful on arbitrary graphs~\cite{ma2018}. We thus use real-time MAPF algorithms that do not guarantee that all agents are successful even if this is possible and evaluate the algorithms by the number of agents that are successful. One could experimentally determine the best MAPF algorithm over a representative set of MAPF instances but one can do better by asking which MAPF algorithm to use when, a question that has been studied more generally in the context of the algorithm selection problem~\cite{Rice:1976}. This decision also has to be made quickly, which suggests using classification algorithms from machine learning to define a fast-to-calculate mapping from features of the MAPF instance to the best performing MAPF algorithm. Consequently, the contribution of our research is applying automated algorithm selection techniques to increase completion rate over using a single algorithm for every problem.

\section{Problem Formulation}\label{sec:problemFormulations}
	
An automatic algorithm selection optimization problem~\cite{Rice:1976} is defined by a tuple $\langle \mathcal{I}, \mathcal{P}, Q \rangle$, where $\mathcal{I} = \{i_1, i_2, \ldots \}$ is a set of problem instances; $\mathcal{P} = \{p_1, p_2, \ldots \}$ is a portfolio of algorithms that can be used to solve each instance $i \in \mathcal{I}$; and $Q: \mathcal{P} \times \mathcal{I} \rightarrow \mathbb{R}$ is a function that returns the quality of the solution found by an algorithm $p \in \mathcal{P}$ when solving problem instance $i \in \mathcal{I}$.
	
A solution to this problem is a mapping $\pi: \mathcal{I} \rightarrow \mathcal{P}$ that maps each problem instance $i \in \mathcal{I}$ to an algorithm $p \in \mathcal{P}$. The quality of a solution $\pi$ is the sum of the qualities of the solutions found by the algorithm prescribed by $\pi$ for each problem instance:
\bea
Q(\pi) = \sum_{i \in \mathcal{I}} Q(\pi(i), i).\label{eq:Q}
\eea
An optimal solution is one that maximizes this value:
\bea
\pi^* = \argmax_{\pi} Q(\pi).
\eea
	
Our set of problem instances $\mathcal{I}$ are multi-agent pathfinding (MAPF) problems on video game maps. We consider a portfolio $\mathcal P$ of MAPF algorithms as candidate algorithms, and we consider \emph{completion rate}, defined as the number of agents that successfully reach their goals within a time limit~\cite{silver:05,Wang:2008}, as our quality metric $Q$\footnote{We equivalently optimize the completion rate averaged over all instances in $\mathcal I$ instead of the cumulative one in Equation~\eqref{eq:Q} above.}. We break ties in $Q$ in favor of algorithms that have short \emph{distances traveled}, defined as the sum of distances travelled over all agents~\cite{silver:05}, followed by algorithms that have small \emph{goal achievement time}, defined as the average wall-clock time it takes for agents to reach their goal from the start of the problem~\cite{silver:05}. Goal achievement times are recalculated when an agent returns to their goal and are undefined if the agent is not in their goal.

\section{MAPF Problem}
\label{pf:mapf}
	
We provide a description of the \emph{multi-agent pathfinding} (MAPF) problem that closely follows the description by \citea{Sigurdson2018}. A MAPF problem is defined by a pair $\left( G, A \right)$, where $G = (N,E,c,h)$ is a undirected weighted graph of nodes $N$ that are connected to each other by edges $E \subset N \times N$. The set of edges includes self-loops, that is, $\forall n \in N:  (n,n) \in E $, which allows all agents to remain on their current node (i.e., wait). We assume that all edge weights $c : E \to \mathbb{R}$ are strictly positive except self-loop edges, which have a weight of $0$. The edge weights are symmetrical $\forall (n,n') \in E : c(n,n') = c(n',n)$.  $A = \{a_1, \ldots, a_n\}$ is a set of NPC agents, where each agent $a_i \in A$ is specified by a pair $(n^i_{\text{start}}, n^i_{\text{goal}})$ that indicates its start node $n^i_{\text{start}}$ and its goal node $n^i_{\text{goal}}$. Each goal is reachable from the start node. The graph is also safely traversable by virtue of being undirected. An estimate of shortest travel distance between any two nodes, the {\em heuristic} $h : N \times N \to \mathbb{R}$, is available to each agent. We use the shorthand $h(n)$ for $h(n,n^i_\text{goal})$ if the goal is understood in the context. An agent may modify $h$ as it sees fits but does not share the modified version with other agents.

In our model, time advances in discrete steps. At time step $t$, each agent $a_i$ occupies a node $n^i \in N$, also referred to as $n_{\text{current}}^i$ when talking about a specific agent's location. When pathfinding, each agent generates a set of moves it plans to execute from its current state. Agents provide these moves to a controller that attempts to have the agent execute its plans one step at a time. Plans are represented as a set of node pairs $P = \{ (n,n') \}$. The pairs represent agent's planned actions (i.e., edge traversals) meaning that when the agent is on node $n$ it intends to go to a neighboring node $n'$ by traversing the edge $(n,n') \in E$.

In the event that the agent's plan is not executable, either because another agent is occupying the node where it wishes to move or because the plan does not have an action planned for the agent's current node, the agent waits in its current node (i.e., traverses the self loop). Agents can traverse edges with the following restrictions: (i) two agents cannot swap locations in a single time step and (ii) each node can be occupied by at most one agent at any time. 
	
As common in video game pathfinding literature, our search graphs in this paper are based on rectangular grids with each grid cell being a single node in the graph~\cite{sturtevant2012benchmarks}. Each grid cell has up to eight immediate neighbors. It is connected to them via cardinal edges of cost $1$ and diagonal edges of cost $\sqrt{2}$.
	
\section{MAPF Algorithms}\label{sec:mapfRelated} 
	
While researchers have proposed a number of algorithms to solve MAPF problems~\cite{wang:11,Shar:15,Wild:13}, in this paper, we focus on using online methods that are better suited for environments, where agents must take actions within a very small amount of time(e.g., video games). Our portfolio is then  comprised of three A*-based algorithms: \emph{Windowed Hierarchical Cooperative A*} (WHCA*)~\cite{silver:05}, \emph{Flow Annotation Replanning} (FAR)~\cite{Wang:2008}, and \emph{Bounded Multi-Agent A*} (BMAA*)~\cite{Sigurdson2018}. We choose these algorithms because they use different strategies to solve MAPF problems and, as a result, can excel on different MAPF problems.

\subsection{Windowed Hierarchical Cooperative A*}

\citea{silver:05} proposed a family of A*-based algorithms for solving MAPF problems: In \emph{Cooperative A*}, each agent runs an A* search in a three dimensional graph ($x$-coordinate, $y$-coordinate, and time) to reach its goal and shares its plan with other agents through reservation tables. Therefore, the agents are able to avoid collisions since each agent knows where all the other agents will be and when they will be there. To improve scalability, \emph{Hierarchical Cooperative A*} uses hierarchical search, where each agent uses the length of the shortest path found in an abstracted state space as a guiding heuristic. Finally, to further improve scalability, \emph{Windowed Hierarchical Cooperative A*} (WHCA*) limits the search depth of each agent to within a window. Once a partial path within the window is found, the agent follows it and searches for the next partial path by shifting the window along the current path. 

\subsection{Flow Annotation Replanning }
	
Like WHCA*, {\em Flow Annotation Replanning} (FAR)~\cite{Wang:2008} also takes account of other agents plans. However, instead of searching for a new path when the current one is blocked, agents in FAR simply wait at their current nodes until they can reserve their next set of moves in a reservation table. FAR also detects deadlocks, where agents would wait on each other indefinitely, and forces them to move away from their current nodes. They then must replan paths back to the node they were forced off of before resuming their original path to their goal.

\begin{figure}[t]
\includegraphics[width=0.95\columnwidth]{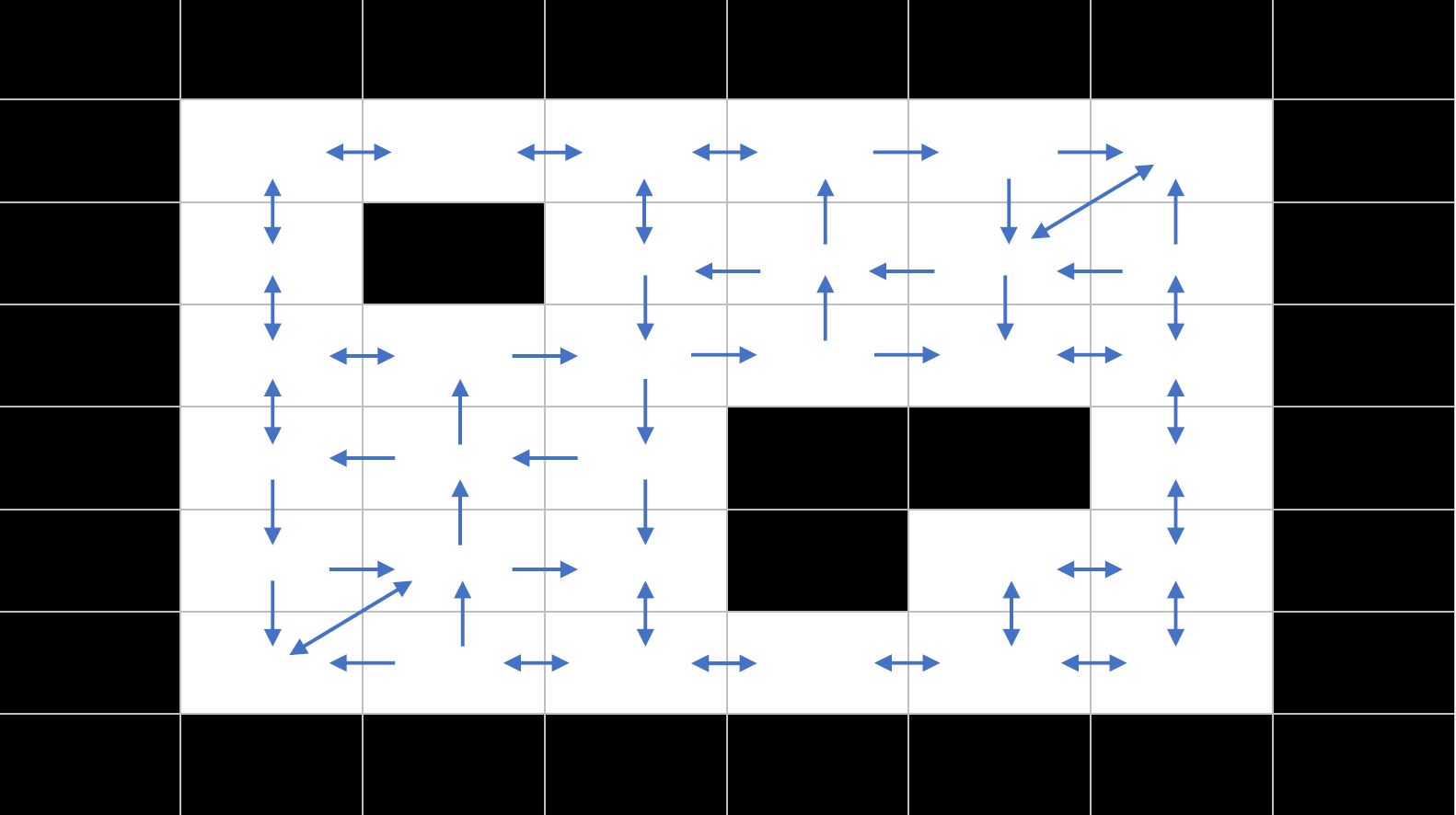}
\caption{Example of Flow Annotations.}
\label{fig:flows}
\end{figure}

To reduce the likelihood of agents blocking each other, FAR annotates at each node with the direction that agents at that node should follow. These annotations combined create ``highways,'' where agents can quickly move from one end of the map to another without ever stopping to wait for other agents. Figure~\ref{fig:flows} shows an example map annotated by FAR which uses the following strategy: It first creates an edge-less annotated graph $G'$ that has the same set of nodes as the original graph $G$. Then, edges are added to $G'$ in alternating directions, that is, even-numbered rows are assigned west-bound edges and odd-numbered rows are assigned east-bound edges. Similarly, even-numbered columns are assigned north-bound edges and odd-numbered columns are assigned south-bound edges. Additional edges are added in special cases. For example, nodes on corridors that are only one-node wide retain their bi-directional connectivity. Self-loops are always retained as agents always have the ability to wait, however self-loops are not considered in the search.

\subsection{Bounded Multi-Agent A*}
	
\emph{Bounded Multi-Agent A*} (BMAA*)~\cite{Sigurdson2018} is based on a \emph{Real-Time Adaptive A*} (RTAA*)~\cite{Koenig:2006}, a well-known \emph{single-agent} real-time heuristic search algorithm. RTAA* runs the following procedures iteratively until the agent reaches its goal: (1) perform a bounded-depth A* search from the agent's current position; (2) update the heuristic values of all nodes in the CLOSED list of that A* search to make them more informed; and (3) move the agent along the partial path returned by that A* search. \citea{Sigurdson2018} extended RTAA* to a \emph{multi-agent} setting, where other agents are treated as (moving) obstacles during the search. Additionally, each agent is able to request other agents that are currently located on its goal cell to vacate. The vacating agent will move to any available neighboring node and resume its regular search procedures from its new location.  

\section{Related Work on Algorithm Selection }\label{sec:aasRelated} 

Performance of planning algorithms can vary substantially based on the problem~\cite{kotthoff2014}. In particular, selecting a heuristic search algorithm specific to the problem can lead to a substantial boost in performance~\cite{bulitkoAiide2016,bulitkoSoCSalife2016}. The latter work showed that the margin for performance improvement over a fixed algorithm grows with the granularity of the selection: Selecting an algorithm per problem instance has a better performance potential than selecting an algorithm per group of problem instances (e.g., problem instances can be grouped based on maps in video games). Follow-up research exploited the potential for performance improvement through machine learning techniques by mapping a problem instance to the best algorithm for that problem instance~\cite{Sigurdson2017}. This work however was limited to the single agent domain and does not provide insight in to how the method would perform in the more realistic multi-agent environment.

Traditionally, machine learning techniques require one to intelligently define input features in order to be successful. However, more recently, deep learning techniques have been able to automatically extract features which allows one to provide low-level problem descriptions. Remarkably, even mapping ASCII codes of a textual problem description in SAT and CSP problems to grey-scale pixel values in a square image provided sufficient for deep convolutional neural networks~\cite{Loreggia:2016:DLA:3015812.3016001}. In pathfinding problems, the encoding can be even simpler since the map itself is naturally represented by a two-dimensional image~\cite{Sigurdson2017}. In this paper, we continue the recent line of work and adapt the latter approach from single-agent to multi-agent pathfinding.

\section{Our Approach}\label{sec:approach}

Before solving our algorithm selection problem defined in Section~\ref{sec:problemFormulations}, there are two key design decisions that must be made: (1) What algorithms to include in the portfolio $\mathcal{P}$ of algorithms? (2) How should problem instances in $\mathcal{I}$ be represented as an input to the selection algorithm? 

\vspace{-\bigskipamount} \paragraph{Portfolio Algorithms:} Ideally, the selection of algorithms in the portfolio should be sufficiently diverse so that for each possible problem instance, there exists at least one algorithm in the portfolio that does well on that problem instance. Larger portfolios, however, may slow down the learning process as well as result in lower performance due to errors when selecting algorithms. Therefore, in this paper, we consider a relatively small but diverse set of algorithms: Windowed Hierarchical Cooperative A* (WHCA*), Flow Annotation Replanning (FAR), and Bounded Multi-Agent A* (BMAA*) as described in Section~\ref{sec:mapfRelated}. We choose BMAA* because we anticipate that it will do well in problem instances where well-informed heuristics are available. In problem instances where heuristics are more substantially misleading, it performs poorly as it runs only bounded-depth searches to find partial paths for the agents, which may be in the wrong direction. Conversely, FAR computes complete paths, but has a limited gridlock breaking procedure that can lead to failure in particularly congested problems. Finally, we choose WHCA* as it can solve particularly tricky problems as it communicates agents plans and takes them into account through a windowed cooperative search. Doing so, however, is computationally expensive and is not always necessary.

\vspace{-\bigskipamount}\paragraph{Input Abstraction:} Videogames will frequently have to solve MAPF problems repeatedly on the same map. Similarities likely extend to patterns in their start and goal locations. It may be desirable for agents to run a specific MAPF algorithm for a given problem. In video games, especially, it is reasonable to assume that the game engine may choose different MAPF algorithms for different games, maps, or even problem instances. Different algorithms may be used within the same map based on the agent's type and priority or even the number of agents. Motivated by this observation, we use a MAPF problem specification with both the map and each agent's start and goal location available. 

\begin{figure}[t]
	\center
	\includegraphics[width=\columnwidth]{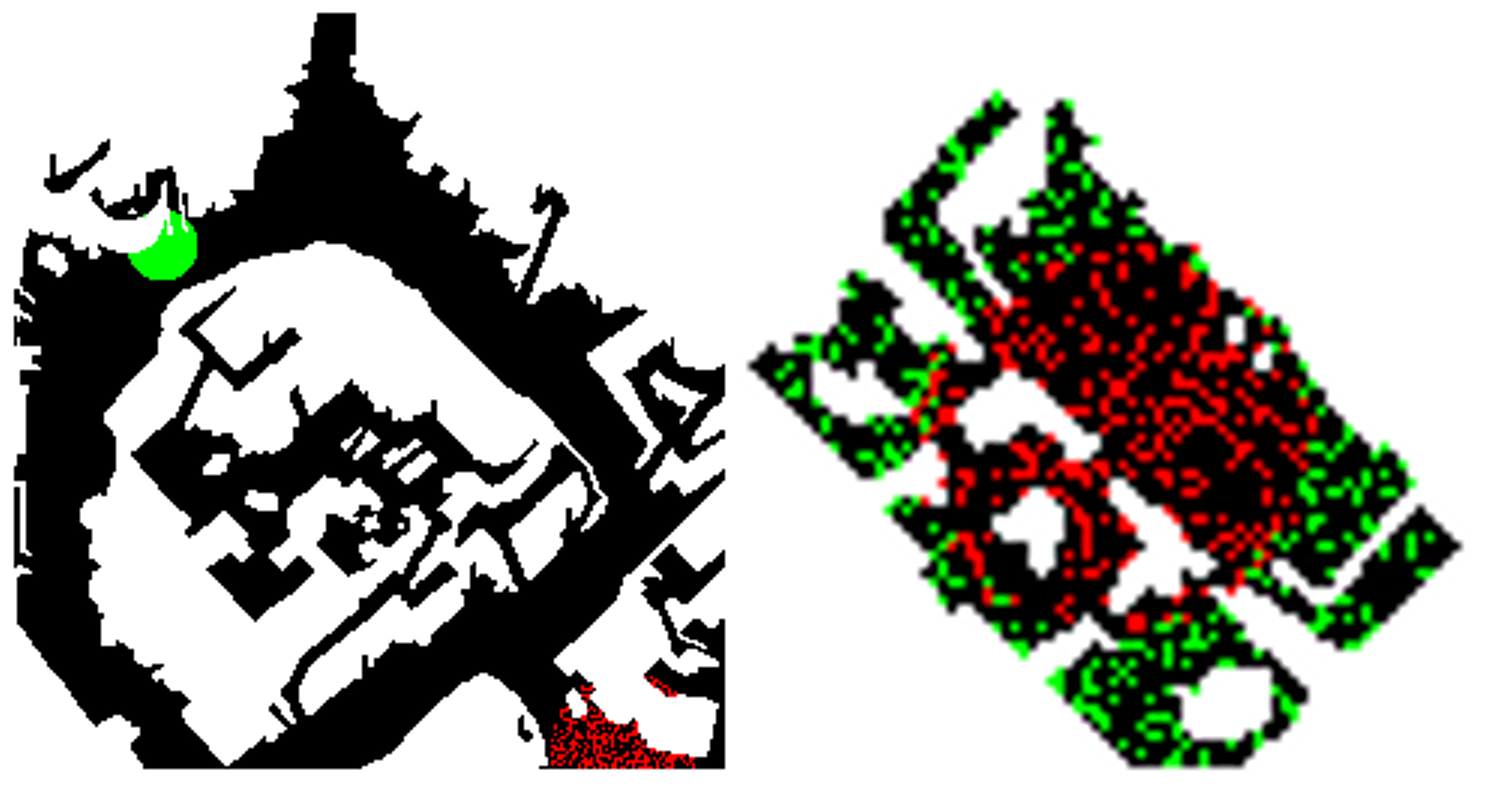}
	\caption{Two sample MAPF problem types with start (green pixels) and goals (red pixels): tight to wide (left) and outside in (right).}
	\label{fig:input}
\end{figure}

\begin{table*}[!htbp]
	\centering
	\caption{Algorithm performance on all problems.}
	\label{tab:results}
	\begin{tabular}{lccccc}
		& Completion Rate(\%)  & & Distance  & & Goal Achievement Time (s) \\ 
		\cline{2-2} \cline{4-4} \cline{6-6}
		\\
		$\pi^*$   & \textbf{$80.8 \pm 0.8$}  &   & $283.1 \pm 7.5$    &   & $15.5 \pm 0.2$              \\
		$\pi$       & $76.6 \pm 1.2$  &   & $261.0 \pm 8.8$    &   & $16.2 \pm 0.4$          \\
		BMAA*     & $65.7 \pm 0.7$  &   & $465.7 \pm 5.9$    &   & \textbf{$14.4 \pm 0.2$}          \\
		FAR       & $66.1 \pm 1.0$  &   & $405.7 \pm 10.1$   &   & $15.9 \pm 0.2$		     \\
		WHCA*     & $54.6 \pm 1.1$  &   & \textbf{$88.3  \pm 1.7$}    &   & $21.7 \pm 0.2$          \\
		Worst     & $44.3 \pm 0.7$  &   & $328.6 \pm 11.0$   &   & $19.7 \pm 0.2$
	\end{tabular}
\end{table*}
\begin{table}[t]
	\centering
	\caption{Completion rate(\%) by problem type.}
	\label{tab:problem}
	\begin{tabular}{lccc}
		
		Problem Type   & BMAA*         & FAR           & WHCA* \\ \hline
		\\
		Random         & $\textbf{79.4}$ & $75.3$		   & $68.9$  \\
		Cross sides    & $69.4 $         & $\textbf{83.7}$ & $56.4$  \\
		Swap sides     & $\textbf{64.4}$ & $48.5$          & $34.5$  \\
		Inside out     & $\textbf{76.7}$ & $75.3$          & $59.9$  \\
		Outside in     & $\textbf{73.3}$ & $72.3$          & $60.7$  \\
		Tight to tight & $33.8$          & $\textbf{37.8}$ & $36.7$  \\
		Tight to wide  & $59.2$          & $\textbf{71.0}$ & $54.9$  

	\end{tabular}
\end{table}

\vspace{-\bigskipamount}\paragraph{Solution Approach:} Our solution approach is motivated by \citea{Sigurdson2017}, who tackled the algorithm-selection problem for single-agent pathfinding. Similar to their approach, we treat the MAPF algorithm-selection problem as an image classification problem. Thus, we classify each image (a representation of a MAPF problem instance) using a fixed set of possible labels (algorithms from the portfolio). We use AlexNet~\cite{krizhevsky2012imagenet}, a \emph{Convolutional Neural Network} (CNN), to solve the image classification problem. We choose a deep learning approach with the anticipation that it will be able to learn to recognize important features of the problem (e.g., topologies of the map, distribution of agents, etc.) and exploit them automatically~\cite{Sigurdson2017}. We choose AlexNet as it is a common, readily-available CNN, which requires no additional engineering. AlexNet is available with PyTorch, Cognitive Neural Tool Kit, TensorFlow, MATLAB and other deep learning frameworks~\cite{AlexnetCNTK,AlexnetPyTorch,AlexnetTensor,vedaldi15matconvnet}.  
In order to provide MAPF problem instances as inputs to the CNN, we represent each problem instance as an image, where blocked and unblocked nodes are represented by white and black pixels, respectively. A partial agent specification is also provided as part of the input, represented by a single RGB pixel. A start location is indicated with a green pixel and a goal location with a red pixel. The mapping between a specific agent's start and goal is anonymized as there is no distinction between which goal belongs to which agent. We resize the image to be equal to that of the network input ($227 \times 227$ pixels) Figure~\ref{fig:input}.

Finally, to generate the dataset to train our CNN, we used publicly available game  maps~\cite{sturtevant2012benchmarks}. Their start and goals were generated as discussed below. For each problem instance in the dataset, we ran each algorithm in our portfolio to determine the best algorithm (i.e., the correct image label) as defined in Section~\ref{sec:problemFormulations}.

\section{MAPF Problem Generation}\label{sec:problemGen}

Rather than using only randomly selected start and goals, we generated different types of MAPF problems. We ensure that start and goals are connected for all the problems generated. These problem types are intended to represent some common video game scenarios. For example, agents \emph{swapping sides} is commonly seen in many strategy games where agents are trying to reach an opposing team's base. \emph{Tight to wide} happens when there is a common spawn location and some general area that the agents are trying to reach. These are not intended to represent all scenarios that occur in games but rather provide a more diverse set of MAPF problem types that may commonly occur in games instead of relying on only a single problem type. 

\begin{itemize}
	\item \emph{Random:} agents are assigned random start and goals.
	\item \emph{Cross sides:} all agents begin on one side (i.e., left, right, top, or bottom) and must traverse to the other side.
	\item \emph{Swap sides:} half the agents start on one side (i.e., left) while the other half start on the opposite side. Their goals are randomly selected in a region on the side opposite of their starting location.
	\item \emph{Inside out:} all agents start near the center of the map and are assigned goals near the outer edges of the map.
	\item \emph{Outside in:} all agents begin near the outer edges of the map and are assigned goals near the center of the map(Figure~\ref{fig:input}).
	\item \emph{Tight to tight:} all agents start as close together as possible and are assigned goals that are as close as possible elsewhere on the map.
	\item \emph{Tight to wide:} all agents start as close together as possible and are assigned goals that are spread out in the same general area of the map(Figure~\ref{fig:input}).

\end{itemize}

\section{Empirical Evaluation}\label{sec:emp}

We modified AlexNet by changing its output layer to match the number of algorithms in our portfolio before training on our dataset of image-label pairs. We use MATLAB's implementation of AlexNet, which is pretrained using the ImageNet dataset~\cite{imagenet}. We use a random $70\%$ of the data for training and the other $30\%$ for testing. All of the values are reported for the test set.

We compare our \emph{automatic algorithm selection} $\pi$ approach against each algorithm in our portfolio (i.e., WHCA*, FAR, and BMAA*), $\pi^*$, which always selects the best algorithm in our portfolio for each problem instance, and a \emph{worst selector}, which always selects the worst algorithm. We set the parameters for the algorithms in our portfolio to the following: the size of reservation tables used for FAR is $3$; the window size of WHCA* is $8$, the lookahead value of BMAA* is $32$ and BMAA* has flow annotations turned off. We chose these values as they performed well in our initial testing. All algorithms were implemented in C\#.

We use the $20$ video-game maps from \emph{Baulder's Gate II} available in the MovingAI benchmark~\cite{sturtevant2012benchmarks}. This included the $10$ largest maps from the game used to originally evaluate FAR~\cite{Wang:2008} and additional $10$ more small and mid sized maps. We included more maps to get a more diverse benchmark ranging from $564$ traversable states to $51586$ states. We fixed the number of agents to $300$ for each MAPF problem. For each map we created $20$ MAPF problems for each of the problem types defined in Section~\ref{sec:problemGen}. Agents not at their goal at the time limit have their goal achievement time artificially set to the $30$ second time limit.

\begin{figure}[t]
\begin{center}
\includegraphics[width=\columnwidth]{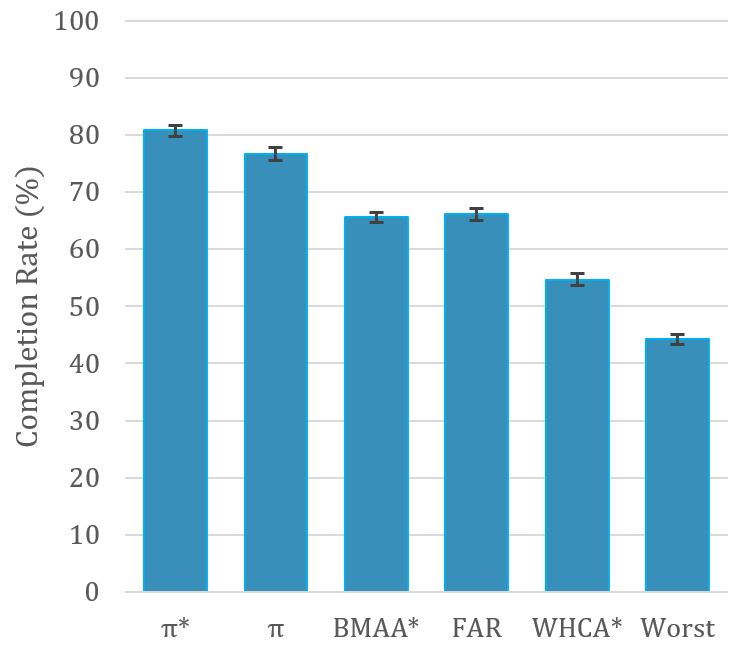}
\end{center}
\vspace{-0.3cm}

\caption{Completion Rate \% for the test data averaged over 10 splits of the data.}
\label{fig:resultsimg}
\end{figure}

Figure~\ref{fig:resultsimg} shows completion rate averaged over $10$ random splits of training and test data to reduce the probability of a favourable split. We provide more details in Table~\ref{tab:results}. We report the performance of the algorithms using our primary completion rate metric, as well as our distances travelled and goal achievement time tie-breaking metrics.

A perfect selector, which uses the best algorithm, would result in a $80.8\%$ completion rate(Table~\ref{tab:results}). Our approach achieves a completion rate of $76.6\%$, which compares favorably to the best single algorithm (FAR), which has a completion rate of $66.1\%$. FAR is followed by BMAA* with a completion rate of $65.7\%$ and WHCA* with a completion rate of $54.6\%$. If one were to consistently select the worst performing algorithm, the completion rate would drop to $44.3\%$.

On the distance traveled metric, which breaks completion rate ties, $\pi$ significantly improves upon BMAA* and FAR with a $44\%$ (from $465.7$ to $260.9$) and $36\%$ ($405.7$ to $260.9$) reduction, respectively. However, the best single algorithm on this metric is WHCA* with $88.3$. Our distance metric is recorded for all agents and not just agents who reach their goal. In other words, an algorithm that never moves the agents would have the smallest distance despite not completing the task. As distance was only for breaking ties in our model, this was not of great concern.

On the goal achievement time metric, BMAA* is better than $\pi$ with a $11.1\%$ (from $16.2s$ to $14.4s$) improvement. On this metric, BMAA* is followed up by WHCA* with a goal achievement time of $15.9s$ and FAR with $21.7s$. The reason why both WHCA* and FAR are slower despite the agents traveling smaller distances than BMAA* is that many of the agents end up waiting for other agents due to their use of reservation tables. WHCA* also performs a more expensive three-dimensional search. 

On average, over the $10$ splits of the training and test data, BMAA*, FAR, and WHCA* was the best choice for $283.5$, $258.9$, and $296.6$ problems, respectively. The neural network predicted BMAA*, FAR, and WHCA* to be the best choice $269.0$, $249.4$, and $321.6$ problems, respectively. $\pi$ used the correct algorithm $72.9\%$ of the time.

WHCA* is often the best choice despite low completion rate across all problem types. WHCA* distance traveled makes it the best choice on problems where all three algorithms achieve $100\%$ completion rate. This leads to its frequent choosing despite its completion rate.

\section{Current Shortcomings and Future Work}
\label{sec:futurework}

Our current approach has been tested on a limited number of maps and configurations. Future work will extend the evaluation to more realistic MAPF problems by mining start and goal locations from actual game logs, to identify if there is significant pathfinding patterns within games for algorithm selection to exploit. We will also consider adding more algorithms to the portfolio as well as assigning algorithms to agents on a per-agent-cluster or a per-agent bases. 

\section{Conclusions}\label{sec:conclusions}

We demonstrated that using an off-the-shelf deep neural network to automatically select multi-agent pathfinding (MAPF) algorithms from a portfolio can improve the performance over the individual algorithms in that portfolio. This approach is promising since it does not require designing new MAPF algorithms. Furthermore, with deep learning, human designers do not even have to handcraft a set of features to describe MAPF problem instances. As a result, this process is accessible to a broad range of game developers. Similar ideas were recently explored for SAT/CSP solver selection~\cite{Loreggia:2016:DLA:3015812.3016001}. However, our approach reduces the engineering effort required by using an off-the-shelf deep neural network (AlexNet) in place of a custom-built CNN.


\bibliographystyle{aaai}
\bibliography{bibliography}

\end{document}